# Dynamic Network Models for Forecasting

Paul Dagum*   Adam Galper   Eric Horvitz*
Section on Medical Informatics
Stanford University School of Medicine
Stanford, California 94305

## Abstract

We have developed a probabilistic forecasting methodology through a synthesis of belief-network models and classical time-series analysis. We present the *dynamic network model* (DNM) and describe methods for constructing, refining, and performing inference with this representation of temporal probabilistic knowledge. The DNM representation extends static belief-network models to more general dynamic forecasting models by integrating and iteratively refining contemporaneous and time-lagged dependencies. We discuss key concepts in terms of a model for forecasting U.S. car sales in Japan.

## 1 INTRODUCTION

Temporal projection and forecasting play a critical role in real-world decision making. Consider the problem of forecasting U.S. car sales in Japan six months, one year, and three years from now. Such forecasts typically are crucial in the planning and execution of corporate and political strategies. A U.S. automobile manufacturer will pay close attention to the predicted Japanese demand for U.S. cars, especially when resources must be allocated for further market penetration. Meanwhile, the U.S. government may adjust its economic and foreign policies to reflect the anticipated penetration of American products in previously inaccessible markets.

Forecasting models are dominated by uncertainty because salient, observable variables define only a small subset of relevant variables; unmodeled influences can lead to unexpected consequences in a dynamic process. Of course, measurement and other errors in the observations, functional relationships among variables, and missing data introduce additional uncertainty.

*Also at the Palo Alto Laboratory, Rockwell International Science Center, 444 High Street, Palo Alto, California 94301.

Statisticians have developed a set of techniques known as *time-series analysis* for forecasting the future values of variables. A *time series* is a set of observations made sequentially over time. Investigators have sought to develop time-series methods for generating or inducing stochastic models, which describe temporal dependencies among successive observations in a time series.

A mature set of probabilistic time-series analysis methods has been developed and applied to a wide range of problems [20]. However, classical time-series methodologies do not provide an expressive representation for capturing the probabilistic dependencies and the nonlinearities of real-world processes. Statisticians have wrestled complex problems into relatively simple parameterized models that can be solved with the traditional methods. Until recently, there has been relatively little interaction between the statisticians interested in time-series analysis and computer-scientists studying the representation of uncertain knowledge with belief networks.

Figure 1 depicts a simple belief-network representation of the interactions between salient variables relevant to the problem of forecasting U.S. car sales in Japan. Briefly, the model states that Japanese demand for U.S. cars depends on the price of these cars and influences the supply of the cars. The price and supply of the U.S. cars is influenced by the efficiency of U.S. manufacturers. Notice that this model captures purely contemporaneous dependencies; that is, the model represents relationships within a fixed time frame. We refer to this model as the *static network*.

In this paper, we address the following problem: Given a static belief network, such as the one depicted in Figure 1, how can we make normative forecasts of the future values of variables? We answer this question by providing an expressive forecasting methodology based on the integration of classical time-series analysis with belief-network representation and inference techniques. Our synthesis has two immediate benefits. First, by casting probabilistic time-series analyses as temporal belief-network problems, we can introduce general dependency models that capture richer, and more realistic, models of dynamic dependencies—as well as the more traditional static or *contemporaneous* belief-network dependencies. Second, we can apply belief-network inference algorithms to the temporal models to do forecasting. The richer models and associated computational methods allow us to move beyond such rigid classical assumptions as linearity in the relationships among variables and normality of their probability distributions.

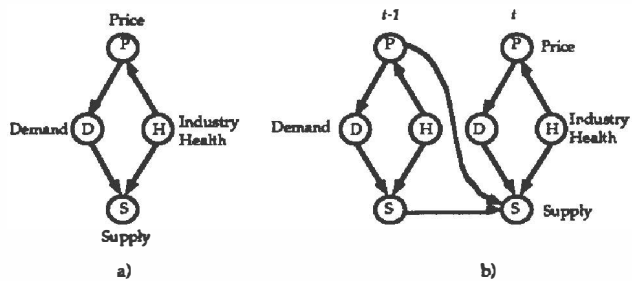

Figure 1: (a) A static network depicting contemporaneous dependencies between economic variables. (b) CARSALES: A DNM for forecasting U.S. car sales in Japan.

In Sections 2 and 5, we discuss methods for generating a forecasting model known as a *dynamic network model* (DNM) from a static network—for example, Figure 1b. Section 3 describes how uncontrolled, unknown exogenous influences can render the forecasting model obsolete unless the model is parameterized; an update method described in Section 3.2 can quickly adapt the model to observed trends. In Section 4, we present procedures for forecasting the future value of a model variable, based on knowledge about the time-lagged values of itself and/or of values of related variables. We conclude with a discussion of the advantages of the DNM approach, and by highlighting representation and inference problems that remain to be solved.

## 2 DYNAMIC NETWORK MODELS

A dynamic model can be constructed from a set of building blocks that capture the instantaneous relationships between domain variables, together with a set of temporal dependencies that capture the dynamic behavior of the domain variables. The building block of a DNM is the traditional, static belief network. We extend the static belief network displayed in Figure 1 to a DNM forecasting model by introducing relevant temporal dependencies between representations of the static network at different times. A DNM for reasoning about future U.S. car sales in Japan is depicted in Figure 1b. Nodes represent states of the domain variables at different times. In this case, the interval between time points is defined to be one month. We refer to this DNM as the CARSALES model.

As highlighted in Figure 1b, we distinguish between two types of dependencies in a DNM. *Contemporaneous dependencies* refer to arcs between nodes that represent variables within the same time period. *Noncontemporaneous dependencies* refer to arcs between nodes that represent variables at different times. Thus, the dependency between *corporate health* and *supply* is contemporaneous, whereas the dependency between this month's *price* and next month's *supply* is non-contemporaenous. The DNM representation allows us to assert and to selectively weaken independence statements about temporal locality by learning noncontemporaneous dependencies and modulating their strength.

Thus far, the specification of the structure of the simple DNM has adhered to traditional belief-network methodology. Our point of departure from the established paradigm is in the specification of the conditional probabilities of the model. As time evolves, so do the potential influences of a plethora of unmodeled exogenous forces that affect the state of the system. The structure and conditional probabilities that define a belief network at one time become obsolete with the passage of time. Discrepancies between new observations and the values predicted by the outdated model attest to a model's inaccuracy and inability to make reliable forecasts. To avoid the progressive deterioration of the model, an intelligent forecasting system must update the conditional probabilities, and also the structure when appropriate, as new evidence arrives.

In the CARSALES model, exogenous influences that may have a significant effect on the Japanese demand for U.S. cars include, for example, new U.S. trade policies or a recent upset in Japanese-U.S. diplomatic relations. In general, these influences are extremely difficult to model adequately, and their inclusion in the model, if done inappropriately, often leads to systematic errors of prediction. Furthermore, important exogenous events may be unknown to the model builder. For example, an undisclosed pending merger-and-acquisition by a Japanese car manufacturer might manifest as a decrease in the corporate health of the U.S. car industry, in spite of the seemingly unchanged economy. An unfortunate presidential blunder during a state dinner, left unpublicized, may also lead to dire forecasts of model variables.

Such sensitivity to unmodeled variables highlights the value of developing a means for updating specific contemporaneous and noncontemporaneous relations that are assessed from a domain expert, as time-series data becomes available. Any probabilistic model that claims to forecast must possess a method of adaptively integrating historical information with current estimates of domain variables. In Section 3.2, we discuss how this requirement imposes the principle of maximum likelihood for the update of conditional probabilities.

# 3  BUILDING AND REFINING A DNM

To build a DNM, an expert specifies the key contemporaneous variables and dependencies. Then, the expert specifies key temporal dependencies among the variables, drawing dependencies from earlier states of the world to variables in the contemporaneous belief network.

Methods for dynamically updating a model specification as new evidence becomes available are critical to techniques for forecasting the future states of complex systems. We simplify the task of updating the DNM model specification by the assumption that the model structure—that is, the set of contemporaneous and noncontemporaneous dependencies—remains invariant over time, and only the set of conditional probabilities need to be updated as time evolves. In other words, we assume that, although variations in unmodeled exogenous forces will inevitably affect the strength of the dependencies, changes in the exogenous environment of the model do not introduce new dependencies or nullify existing dependencies. For many systems, this assumption is valid. An example that invalidates the assumption, in the context of a simple supply-demand model, occurs when an essential commodity—that is, one for which demand is virtually fixed—is supplied by a monopolistic power. Monopolistic control of a commodity arises from a cartel between commercial enterprises designed to fix prices regardless of the demand.

To design a DNM that dynamically updates conditional probabilities when new evidence becomes available, we need to address two crucial problems. First, we must specify conditional probabilities that are amenable to incremental adjustments which reflect changes in exogenous forces—the *assessment task*. Then, we must specify how, given new observations, we adjust the conditional probabilities without introducing biases and with minimization of the expected error in the forecasts—the *update task*.

## 3.1  Assessing DNM Probabilities

To assess the conditional probabilities, we employ an approach that draws on the best elements of expert-assessment and parameter-estimation techniques. The parametric decomposition is chosen by an expert in anticipation of the key components of the conditional probabilities that need to be adjusted to reflect sudden changes in the unmodeled exogenous environment. The expert strives to attain a balance between an expressive and a parsimonious parameterization. The use of a conditional probability table for each node, with each table entry treated as an updatable parameter, is an expressive parameterization, yet complex parameterization. The complexity of the assessment procedure virtually precludes making unbiased adjustments of the conditional probabilities. In Section 5 we discuss two parameterization models for the conditional probabilities. These models are motivated by the methods used currently by time-series analysts. Validity of the models has been substantiated by the overwhelming success witnessed by three decades of time-series analysis using these models.

To obtain probabilities for the CARSALES model, we first assess from an expert the functional form of the conditional probabilities. The expert may be able to specify qualitatively the form of the conditional probabilities by choosing to focus on the key determinants of the distribution that are known to be time invariant. For example, the supply of U.S. manufactured cars will decrease invariably in direct proportion to the Japanese demand. Although last month's manufacture price and the U.S. car industry health will modulate the absolute effect, the *supply* node's conditional probability should still reflect the putative relationship between supply and demand.

Once the appropriate parameterization has been defined, the conditional probabilities can be refined to reflect the evidence at hand. Such refinement is achieved either through the estimation of parameter values from data using maximum-likelihood methods or through user specification. In either case, the advantage of this approach is a parsimonious and adaptive representation of the conditional probabilities.

## 3.2  Updating the Model with Data

In time-series analyses, dynamic models adapt to changes in system behavior through the reestimation of model parameters when new observations are made. The process of *updating* CARSALES is the iterative process by which new observations are used to update the maximum-likelihood estimates of the likelihood weighting coefficients in Equation 1. A *model update* of CARSALES is an adaptive learning procedure in which the underlying model is changed; it should be distinguished from a *belief update*. In the model update, we use new evidence to update our prior belief in the model specification; in contrast, in a belief update, new evidence changes our belief in a proposition inferred from the belief network. Both methods use maximum likelihood to update the respective beliefs. Once the parameters are updated, the conditional-probability distributions of the model are reevaluated.

## 3.3  Beyond Two-State Models

The specification of the CARSALES DNM tacitly assumes that maintaining observations from two distinct times is sufficient to provide reliable forecasts. This simplistic model is intended only to highlight the essential features of DNMs. More realistic models for forecasting car sales in Japan might include an array of domain variables and dependencies intended to capture important exogenous influences. Included among these dependencies would be *lag* dependencies—that is, dependencies between nonadjacent time points. These dependencies can model the lagged effects of a policy; for example, federal tax breaks for the car industry would be expected to have an effect on industry health only in the subsequent fiscal year. Lag dependencies are critical in modeling the cyclical behavior of domain variables. The Japanese demand can be expected to display seasonality—that is, it should follow closely the seasonal behavior of unemployment. If we are interested in the seasonal behavior of Japanese demand, we would structure CARSALES differently.

### 3.4 Diagnostic Checks

If the intent of building a DNM is to provide a means for making optimal forecasts of future values of endogenous variables, observed values should be in the immediate neighborhood of predicted values. The *residuals* of a DNM at time $t$ denote the difference between the one-step-ahead forecasts and the values observed at time $t$. If the DNM is correctly assessed by the expert, the residuals should be distributed normally and independently with mean zero. A plot of the residual sample autocorrelations should not reveal the presence of serial correlation. A serial correlation indicates that there is some systematic aspect of the behavior of the system that is not being detected by the model. Thus, diagnostic checks assess the adequacy of the model and can serve to suggest appropriate modifications.

## 4  FORECASTING: INFERENCE WITH A DNM

Ongoing implementations of complex DNMs include a sleep apnea model and a model for surgical intensive-care ventilator management. Validation of the performance of these models is underway. Here, we shall highlight key phases of updating and forecasting with DNMs with analytical results derived from the CARSALES DNM.

To forecast at time $t$ the $t+1$ values of the nodes in CARSALES, we *scroll* the model one time slice into the future. Figure 2 depicts the process of scrolling the model. The scrolled model uses the same values estimated at time $t$ for the likelihood weights combining the contemporaneous and the noncontemporaneous influences on supply. Thus, the model implicitly reflects the effects of unmodeled exogenous forces on the level of the forecasts, thus, increasing the forecast reliability. In general, to project $l$ time points into the future, a DNM at time $t$ sequentially steps through the time points $t+1, ..., t+l$. Thus, a complete profile of the time series over the time interval from $t$ to $t+l$ is constructed in the process of forecasting.

Once the projection model at time $t+l$ has been identified, the forecasted values of the endogenous variables at time $t+l$ are computed by a probabilistic inference algorithm. For typically complex applications, the size and topology of the DNM may prohibit tractable exact computation of inferences. In such cases, stochastic-simulation algorithms designed specifically to approximate inference probabilities in large belief networks can be employed [15, 14, 11, 16, 9, 17, 4, 6, 7].

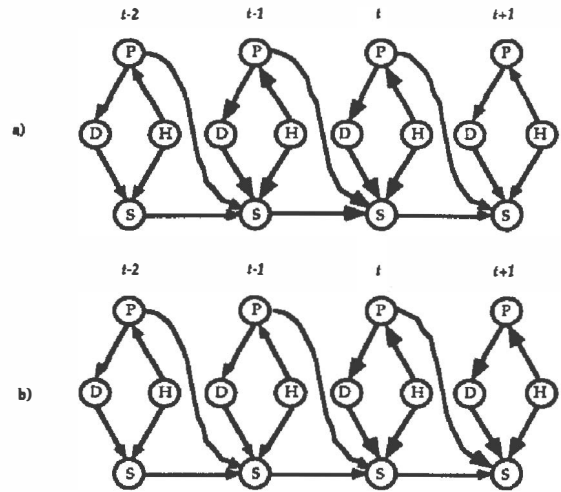

Figure 2: Forecasting with CARSALES. In going from figure (a) to figure (b), the model is scrolled on time slice into the future. The forecasted values at time $t+1$ are computed from the model in figure (b), while preserving the likelihood weights computed in (a) and the set of observations made in (a) for nodes at time $t$.

## 5  SPECIAL DNMS

We can simplify the assessment of conditional probabilities for DNMs by employing special parameterized functional forms. We shall focus on two simple parametric decompositions used commonly by time-series analysts: the *additive* and the *multiplicative* decompositions. The additive decomposition is used commonly in time-series analysis for integrating predictions based on current observations with predictions based on historical observations. Additive decompositions are an integral aspect of models that purport to forecast future values of time series, and they appear in the form of the Kalman filter in state-space models, and in the conditional sum of squares in ARIMA models [10]. The multiplicative decomposition is used commonly to model log-linear systems in engineering applications.

Both decompositions employ *likelihood weights*, which provide a language for assigning measures of reliability to information about different times. With this approach, we consider the probabilistic dependencies from contemporaneous sets of variables and from variables at different points in the past as providing independent sources of information. These measures are used to weight the contribution of the contemporane-ous and noncontemporaneous dependencies differently. The sum of the predictions, each weighted by its likelihood, gives the final prediction. The use of likelihood weighting allows an expert to specify the weight of the past versus the present with ease. Such parameters in the functional forms also allow the distributions to be tuned adaptively as dictated by a set of current observations.

In the CARSALES model, the supply at time $t$, based on information at time $t$, depends on the demand and industry health at time $t$, denoted by $Q[s_t|d_t, h_t]$. The supply at time $t$, based on information prior to time $t$, depends on the price and supply at time $t-1$, denoted by $R[s_t|p_{t-1}, s_{t-1}]$. Let $\alpha$ denote the likelihood that the supply predicted from the information prior to time $t$ is correct; therefore, $1 - \alpha$ denotes the likelihood that the supply predicted from information at time $t$ is correct. In the additive decomposition, the prediction of supply is given by

$$\Pr[s_t|d_t, h_t, p_{t-1}, s_{t-1}; \alpha] = \alpha Q[s_t|d_t, h_t] \\ + (1-\alpha) R[s_t|p_{t-1}, s_{t-1}].$$

In the multiplicative decomposition, the prediction is

$$\Pr[s_t|d_t, h_t, p_{t-1}, s_{t-1}; \alpha] = \mathcal{N} Q[s_t|d_t, h_t]^\alpha \\ \times R[s_t|p_{t-1}, s_{t-1}]^{1-\alpha}$$

where $\mathcal{N}$ is a constant that normalizes the probability to unity.

We proceed to study the adaptive behavior of the additive decomposition given new evidence. We focus on the behavior of the CARSALES model. Let $s_i$, $i \leq t$, denote the values of the supply nodes observed for time points up to, and including, time point $t$, and similarly, let $\xi_i$, $i \leq t$, denote the set of values observed for nodes $d_i, h_i, p_i$. We are interested in evaluating the conditional likelihood function $L[s_t, s_{t-1}|\xi_t, \xi_{t-1}, s_{t-2}, p_{t-2}; \alpha]$, predicted by our model CARSALES, of observing $s_{t-1}$ and $s_t$. From Equation 1 the conditional probability for the supply node depends on the chosen value of the likelihood weight $\alpha$, and we expect $L$ to vary accordingly. If $\alpha^*$ maximizes $L$, then $\alpha^*$ is the maximum-likelihood estimator of $\alpha$, and it is optimal in the sense that it is an unbiased estimator of $\alpha$ that achieves the Cramér-Rao lower bound on the variance for all possible unbiased estimators. Accordingly, predictions made by CARSALES using maximum-likelihood estimators for the parameters are optimal over the space of all possible unbiased estimators.

We begin by giving the expression for $L$ in the CARSALES example,

$$L = \Pr[s_t|d_t, h_t, p_t, s_{t-1}; \alpha] \\ \times \Pr[s_{t-1}|d_{t-1}, h_{t-1}, p_{t-1}, s_{t-2}; \alpha].$$

If we substitute for each conditional probability in the preceding expression for $L$ the expression given by the additive decomposition, the resulting equation is a quadratic equation in $\alpha$.

$$\begin{aligned} L = {}& \alpha^2 \delta_t \delta_{t-1} \\ & + \alpha \left[ \delta_t R[s_{t-1}|s_{t-2}, p_{t-2}] + \delta_{t-1} R[s_t|s_{t-1}, p_{t-1}] \right] \\ & + R[s_{t-1}|s_{t-2}, p_{t-2}] R[s_t|s_{t-1}, p_{t-1}] \end{aligned} \quad (1)$$

where, for $i = t-1, t$,

$$\delta_i = Q[s_i|d_i, h_i] - R[s_i|s_{i-1}, p_{i-1}]. \quad (2)$$

Equation 1 is the equation of a parabola with extremum occurring at

$$\alpha_m = -\frac{\delta_t R[s_{t-1}|s_{t-2}, p_{t-2}] + \delta_{t-1} R[s_t|s_{t-1}, p_{t-1}]}{2 \delta_t \delta t - 1} \quad (3)$$

The maximum-likelihood estimator $\alpha^*$ must lie in the interval $[0,1]$. The constraint leads to the following two cases: (1) if $\delta_{t-1} \delta_t > 0$, then the parabola is convex-up, and thus, $\alpha^* = 0$ if $\alpha_m \leq \frac{1}{2}$ and $\alpha^* = 1$ if $\alpha_m > \frac{1}{2}$; (2) $\delta_{t-1} \delta_t \leq 0$, and, either $\alpha_m \leq 0$ and $\alpha^* = 0$, or $0 < \alpha_m \leq 1$ and $\alpha^* = \alpha_m$, or $\alpha_m > 1$ and $\alpha^* = 1$.

In terms of the model structure, a choice of $\alpha^* = 1$ implies that predictions based on prior information should be ignored categorically. This conclusion is consistent with the finding that, if $\delta_{t-1}, \delta_t > 0$, for $i = t-1, t$, the values of $d_i$ and $h_i$ are correlated more strongly with the outcome $s_i$ than are the values of $s_{i-1}$ and $p_{i-1}$. The conclusion follows because *prior* to observing the outcome for supply at time $i$, the probabilities $Q[S_i|d_i, h_i]$ and $R[S_i|s_{i-1}, p_{i-1}]$ are distributions for the node $S_i$ predicted from current information, $d_i$ and $h_i$, and from prior information, $s_{i-1}$ and $p_{i-1}$. Thus, these probabilities are a measure of the correlation of the current and the prior information with the finding—that is, the observed outcome $s_i$. Conversely, a choice of $\alpha^* = 0$ implies that predictions based on current information should be categorically ignored. When $\delta_{t-1}$ and $\delta_t$ are of opposite signs, the appropriate choice of $\alpha^*$ may take on a value intermediate between 0 and 1, implying that the prediction for supply based on maximum-likelihood is a weighted mix of the two predictions.

Once $\alpha$ has been computed—that is, updated to reflect new observations—the model can be used to forecast next month's supply. We scroll CARSALES forward one time point, and compute the marginalized distribution $\Pr[S_{t+1}|\xi]$ in the belief network using probabilistic inference. For large models, or for models used in high stakes time-pressured environments, an approximate algorithm that trades off accuracy of inference for efficiency of computation is preferred.

## 6  NUMERICAL EXAMPLE

Let us consider an example using the additive model for CARSALES. We assume that all nodes are binary—that is, *demand, price, industry health* and *sales* are all either *high* (H) or *low* (L). The conditional probability distributions for $Q[s_t = H|d_t, h_t]$ and $R[s_t = $ $H \mid s_{t-1}, p_{t-1}]$ are given in Figure 3. The remaining conditional probabilities for CARSALES are given in Figure 4.

| $d_t h_t$ | $Q[s_t = H \mid d_t, h_t]$ |
|---|---|
| HH | 0.55 |
| HL | 0.25 |
| LH | 0.60 |
| LL | 0.55 |

| $p_{t-1} s_{t-1}$ | $R[s_t = H \mid p_{t-1} s_{t-1}]$ |
|---|---|
| HH | 0.90 |
| HL | 0.40 |
| LH | 0.40 |
| LL | 0.10 |

Figure 3: The conditional probabilities $Q[s_t = H \mid d_t, p_t]$ and $R[s_t = H \mid s_{t-1}, p_{t-1}]$ in CARSALES. In the example, the conditional probability for the car-sales node is given by the convex combination of the probabilities $Q$ and $R$ with parameter $\alpha$.

| $p_t$ | $\Pr[d_t = H \mid p_t]$ |
|---|---|
| H | 0.25 |
| L | 0.65 |

| $h_t$ | $\Pr[p_t = H \mid h_t]$ |
|---|---|
| H | 0.35 |
| L | 0.80 |

Figure 4: The conditional probabilities $\Pr[d_t = H \mid p_t]$ and $\Pr[p_t = H \mid h_t]$ of CARSALES. The prior probability for the industry health is $\Pr[h_t = H] = 0.85$.

| t | 0 | 1 | 2 | 3 | 4 | 5 | 6 |
|---|---|---|---|---|---|---|---|
| $d_t$ | H | H | H | L | L | L | L |
| $h_t$ | H | H | H | H | H | H | H |
| $p_t$ | H | H | H | H | H | H | L |
| $s_t$ | L | L | L | H | H | H | H |
| $\alpha^*$ | | 0.0 | 0.0 | 1.0 | 0.5 | 0.0 | 0.0 |
| $\overline{s}_{t+1}$ | | 0.40 | 0.40 | 0.56 | 0.73 | 0.90 | 0.90 |

| t | 7 | 8 | 9 | 10 | 11 |
|---|---|---|---|---|---|
| $d_t$ | L | L | L | L | L |
| $h_t$ | H | H | L | L | L |
| $p_t$ | L | L | L | L | L |
| $s_t$ | H | H | L | L | L |
| $\alpha^*$ | 0.5 | 1.0 | 1.0 | 0.0 | 0.0 |
| $\overline{s}_{t+1}$ | 0.48 | 0.56 | 0.56 | 0.10 | 0.10 |

Figure 5: Time-series of data for CARSALES over twelve periods. Values for $\alpha^*$ are maximum-likelihood estimates corresponding to the observed data. For each period, we forecast the probability that the supply will be high in the next period. We denote this forecast by $\overline{s}_{t+1}$, and we compute it by evaluating the inference $\Pr[s_{t+1} = H \mid \xi_t]$ in CARSALES with the $\alpha^*$ computed at time $t$ and where $\xi_t$ denotes all observations made up to and including time $t$.

Figure 5 contains a time-series of observations of the variables of CARSALES made over twelve consecutive periods. Initially, the time-series is assumed to be stationary with data reflecting a healthy U.S. car industry and high-volume sales in Japan. The model is in equilibrium with the time-series, where in this example, the equilibrium is reflected in the value of $\alpha^*$, which takes on the value 0 when equilibrium is reached. Exogenous events that disturb the level of the time-series occur at $t = 3, 6, 9$. CARSALES adapts to the exogenous disturbance with $\alpha^*$ increasing and subsequently decreasing over ensuing periods. The number of periods over which $\alpha^*$ increases is a function of the lagged influences in the model—that is, the noncontemporaneous relations.

At $t = 3$ we observe a sudden drop in the demand which persists through subsequent time periods. The persistence of this drop allows us to rule out noise as the cause, and we can assume that the change in the demand has been precipitated by an exogenous event.

Concomitant with the low demand there is an increase in supply, while both industry health and price remain temporarily unchanged. CARSALES responds to the disturbance by temporarily increasing $\alpha^*$ until a new equilibrium is achieved. The forecasted probabilities for a high supply rise from 0.4 to 0.9 during this period. At $t = 6$, we observe a drop in price which once again disturbs the equilibrium state. The low price continues through remaining periods, and reflects, to some extent, the market forces that result form high supply and low demand which are captured by the supply-demand model CARSALES. However, the probability that the price drops in CARSALES when demand is low and supply and industry health are high, cannot alone account for the persistence of the low price. We anticipate that the price has been exogenously clamped at a low value. Although a single drop in price at $t = 6$ does not affect the value of $\alpha^*$, the consistently low values represent an unlikely circumstance. The CARSALES model responds to this scenario by adjusting $\alpha^*$ during the subsequent two time periods. The forecasted probability for supply made in period 7 drops in anticipation of a drop in supply by period 8, but which is first observed at period 9. At $t = 9$, subject to declining prices and demand, the health of the industry deteriorates, and consequently, supply drops as well. Once again, the disturbance is sensed by CARSALES through the parameter $\alpha^*$, and the forecasts for supply drop even further. In the absence of further changes, by period 11 the model is again at equilibrium.

## 7  RELATED WORK

Most research in temporal reasoning has focused on the representation of time with logical predicates, rather than on dynamic modeling of the state of the world under uncertainty [2, 13, 18]. Several approaches have been proposed to support temporal reasoning using probability theory. To date, most of this research has been hindered by problems with capturing dynamic changes in temporal models as new observations become available. Early attempts to develop probabilistic methods for temporal reasoning have posed static models of dynamic domains, in which exogenous influences are captured in fixed conditional-probability distributions. Such models rely entirely on prior knowledge of the domain, and offer no method for refining the probabilistic dependencies of the models dynamically with new data. DNMs can adapt to exogenous influences by fine tuning their conditional-probability distributions. Such dynamic adaptation can reduce forecasting errors and improve planning and control.

Cooper, et al. [5] propose methods for encoding uncertain temporal relationships under several restrictive assumptions, to allow the exact computation of the joint probability of a hypothesis and the accumulated temporal evidence. Dean and Kanazawa [8] develop a probabilistic model for projection based on a functional (e.g., exponential) decay model of the persistence with time of propositions. Berzuini [3] embeds semi-Markov models in a belief-network representation and uses approximate probabilistic inference to compute the degree of belief in past states and in future states. Tatman shows that a Markov decision process can be encoded in an influence diagram [19]. Kanazawa and Dean [12] apply approximate decision-making processes to these influence diagrams to trade off accuracy of prediction for speed of decision making.

Most of the models developed in research on temporal reasoning have a limited ability to adapt to new observations, and their Markov nature, prevents them from making forecasts that extend beyond Markov simulations. Recent work by Abramson [1] describes a belief-network model that predicts future crude-oil prices given historical evidence. However, he employs classical time-series methods external to the belief-network to reestimate model parameters.

## 8  CONCLUSION

The temporal dependencies and time-based evolution of the states of variables in a dynamic domain limit the applicability of conventional static belief-network models. DNMs extend classical dynamic modeling, by providing an expressive language and platform for ongoing research on probabilistic temporal reasoning. DNMs are ideally suited for forecasting and control in domains for which detailed prior knowledge is available about the dynamic forces and relations at play, but is sufficiently complex to preclude a complete specification. These are the domains that we face in probabilistic-reasoning applications. With the DNM approach, machinery is provided for compensating for the exogenous influences in an unbiased fashion. Our future research on DNMs includes exploring the performance of alternate inference algorithms to solve special DNM topologies, validating the predictive behavior of alternative models, and investigating methods for inducing DNMs from static belief networks by identifying temporal dependencies from time-series data.

## Acknowledgments

This work was supported by the National Science Foundation under grant IRI-9108385 and Rockwell Science Center IR&D funds.